\documentclass[10pt,letterpaper]{article}
\usepackage[top=0.85in,left=2.75in,footskip=0.75in]{geometry}

\usepackage{amsmath,amssymb}

\usepackage{changepage}

\usepackage[utf8x]{inputenc}

\usepackage{textcomp,marvosym}

\usepackage{cite}

\usepackage{nameref,hyperref}

\usepackage{microtype}
\DisableLigatures[f]{encoding = *, family = * }

\usepackage[table]{xcolor}

\usepackage{array}

\newcolumntype{+}{!{\vrule width 2pt}}

\newlength\savedwidth

\raggedright
\usepackage[aboveskip=1pt,labelfont=bf,labelsep=period,justification=raggedright,singlelinecheck=off]{caption}

\makeatletter
\renewcommand{\@biblabel}[1]{\quad#1.}
\makeatother

\usepackage{lastpage,fancyhdr,graphicx}
\usepackage{epstopdf}
\fancyheadoffset[L]{2.25in}
\fancyfootoffset[L]{2.25in}
\begin{document}

\begin{flushleft}
\textsc{research article}
\vspace*{0.2in}
{\Large
\textbf\newline{The natural selection of words: Finding the features of fitness} 
}
\newline
\\
Peter D. Turney\textsuperscript{1*},
Saif M. Mohammad\textsuperscript{2$\dagger$}
\\
\bigskip
\textbf{1} Ronin Institute, Montclair, New Jersey, United States of America
\\
\textbf{2} National Research Council Canada, Ottawa, Ontario, Canada
\\
\bigskip

%
%


* peter.turney@ronininstitute.org

$\dagger$ saif.mohammad@nrc-cnrc.gc.ca

\end{flushleft}
\section*{Abstract}

We introduce a dataset for studying the evolution of words, constructed 
from WordNet and the Google Books Ngram Corpus. The dataset tracks the 
evolution of 4,000 synonym sets ({\em synsets}), containing 9,000 English 
words, from 1800 AD to 2000 AD. We present a supervised learning algorithm 
that is able to predict the future leader of a synset: the word in the 
synset that will have the highest frequency. The algorithm uses features 
based on a word's length, the characters in the word, and the historical 
frequencies of the word. It can predict change of leadership (including 
the identity of the new leader) fifty years in the future, with an F-score 
considerably above random guessing. Analysis of the learned models 
provides insight into the causes of change in the leader of a synset. 
The algorithm confirms observations linguists have made, such as 
the trend to replace the {\em -ise} suffix with {\em -ize}, the rivalry 
between the {\em -ity} and {\em -ness} suffixes, and the struggle between 
economy (shorter words are easier to remember and to write) and clarity 
(longer words are more distinctive and less likely to be confused with
one another). The results indicate that integration of the Google Books 
Ngram Corpus with WordNet has significant potential for improving our 
understanding of how language evolves.

\section*{Introduction}

Words are a basic unit for the expression of meanings, but the mapping
between words and meanings is many-to-many. Many words can have
one meaning (synonymy) and many meanings can be expressed with one
word (polysemy). Generally we have a preference for one word over
another when we select a word from a set of synonyms in order to 
convey a meaning, and generally one sense of a polysemous word is more
likely than the other senses. These preferences are not static; they
evolve over time. In this paper, we present work on improving our 
understanding of the evolution of our preferences for one word
over another in a set of synonyms.

The main resources we use in this work are the Google Books Ngram 
Corpus (GBNC) \cite{Michel2011,Google2012,Google2013} and WordNet 
\cite{Fellbaum1998,WordNet2007}. GBNC provides us with information
about how word frequencies change over time and WordNet allows
us to relate words to their meanings.

GBNC is an extensive collection of word ngrams, ranging from unigrams 
(one word) to five-grams (five consecutive words). The ngrams were 
extracted from millions of digitized books, written in English, 
Chinese, French, German, Hebrew, Spanish, Russian, and Italian 
\cite{Michel2011,Google2012,Google2013}. The books cover the years 
from the 1500s up to 2008. For each ngram and each year, GBNC 
provides the frequency of the given ngram in the given year 
and the number of books containing the given ngram in the given year. 
The ngrams in GBNC have been automatically tagged with part of speech 
information. Our experiments use the full English corpus, 
called {\em English Version 20120701}.

WordNet is a lexical database for English \cite{Fellbaum1998,WordNet2007}.
Similar lexical databases, following the format of WordNet, have been
developed for other languages \cite{Vossen1997,Vossen2001}. Words in 
WordNet are tagged by their parts of speech and by their senses. A 
fundamental concept in WordNet is the {\em synset}, a set of synonymous 
words (words that share a specified meaning).

\begin{figure*}[!ht]
\begin{adjustwidth}{-2.25in}{0in}
\begin{center}
\includegraphics[width=\linewidth]{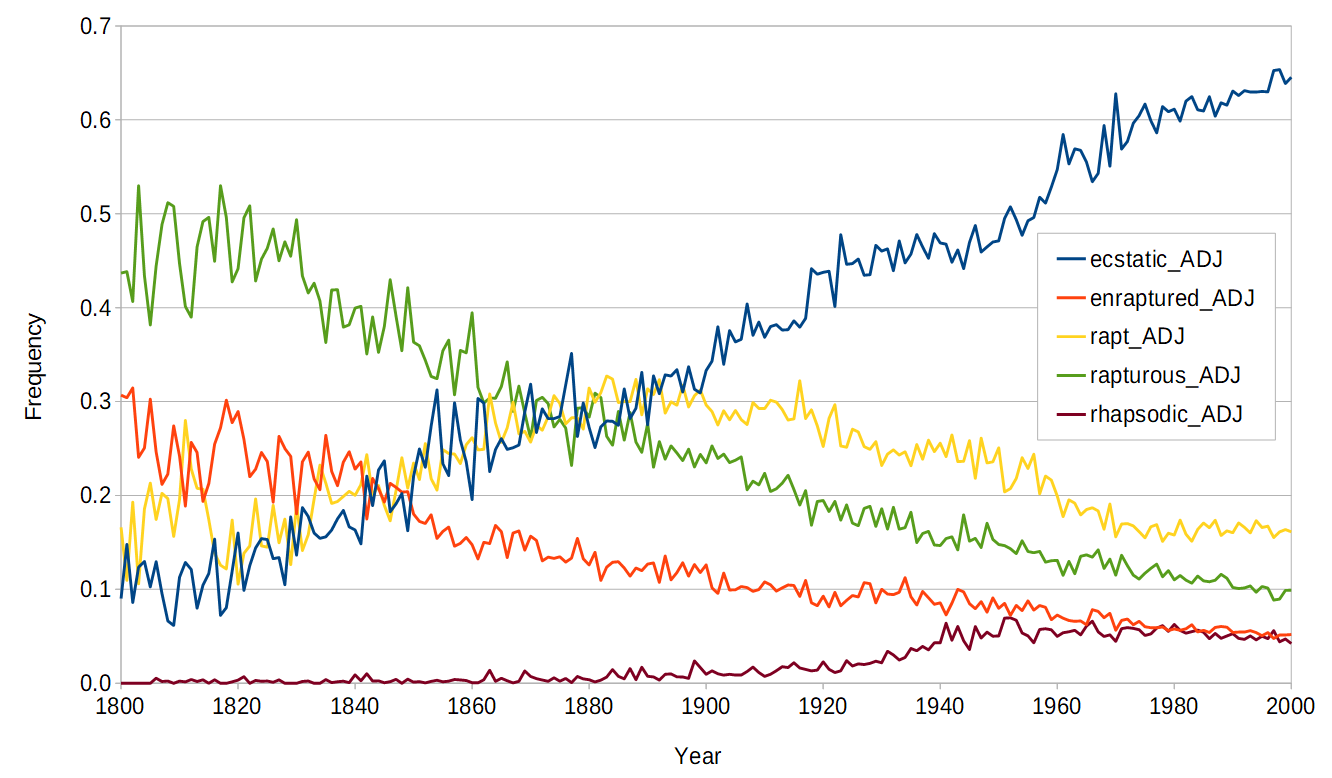}
\end{center}
\caption{\label{figure:rapturous--ecstatic} 
\textbf{The normalized frequencies of the {\em rapturous--ecstatic} 
synset from 1800 AD to 2000 AD.} 
\newline
The sum of the five frequencies for any given year is 1.0. The data has 
not been smoothed, in order to show the level of noise in the trends.
This synset is typical with respect to the shapes of the
curves and the level of noise in the trends, but it is atypical in that 
it contains more words than most of the synsets. Most of the synsets
contain two to three words.}
\end{adjustwidth}
\end{figure*}

According to WordNet, {\em ecstatic}, {\em enraptured}, {\em rapt}, 
{\em rapturous}, and {\em rhapsodic} all belong to the same synset, 
when they are tagged as adjectives ({\em enraptured} could also be 
the past tense of the verb {\em enrapture}). They all mean ``feeling 
great rapture or delight.'' Based on frequency information from GBNC, 
Fig~\ref{figure:rapturous--ecstatic} shows that {\em rapturous} was the 
most popular member of this synset from 1800 AD to about 1870 AD. After 
1870, {\em ecstatic} and {\em rapt} competed for first place. By 1900, 
{\em ecstatic} was the most popular member of the synset, and its lead 
over the competition increased up to the year 2000. For convenience, 
we will refer to this as the {\em rapturous--ecstatic} synset.

Competition among words is analogous to biological evolution 
by natural selection. The leading word in a synset (the word with the
highest frequency) is like the leading species in a genus (the species with
the largest population). The number of tokens of a word in a corpus 
corresponds to the number of individuals of a species in an environment.  

Brandon \cite{Brandon1996} states that the following three 
components are crucial to evolution by natural selection:

\begin{quote}
\begin{em}
\begin{enumerate}
\item Variation: There is (significant) variation in morphological, 
physiological and behavioural traits among members of a species.
\item Heredity: Some traits are heritable so that individuals resemble 
their relations more than they resemble unrelated individuals and, in 
particular, offspring resemble their parents.
\item Differential Fitness: Different variants (or different types of 
organisms) leave different numbers of offspring in immediate or remote 
generations.
\end{enumerate}
\end{em}
\end{quote}

\noindent Godfrey-Smith \cite{Godfrey-Smith2007} lists the same 
three components, calling them {\em conditions for evolution by natural 
selection}.

When a system satisfies these three conditions, we have evolution 
by natural selection. Synsets satisfy the conditions. There is 
{\em variation} in the words in a synset: new words are coined and 
enter a synset, old words gain new meanings and enter a synset. There 
is {\em heredity} in word formation: this heredity is investigated in 
the field of etymology. There is {\em differential fitness}: some 
words become more popular over time and increase in frequency, other 
words become less popular and decline in frequency. Thus we may say 
that synsets evolve by natural selection.

Our focus in this paper is on {\em differential fitness}, also known as 
{\em competition} or {\em selection}. Selection determines which word will 
dominate (with respect to frequency or population) a synset. Here we do not 
attempt to model how new words are formed (variation) or how tokens are 
reproduced with occasional mutations (heredity), although these are 
interesting topics.

A number of recent papers have examined the problem of understanding how words 
change their meanings over time \cite{Mihalcea2012,Mitra2015,Xu2015,Hamilton2016}. 
In contrast, we examine the problem of understanding how {\em meanings} (synsets) 
change their {\em words} over time. Words compete to represent a meaning, just 
as living organisms compete to survive in an environment. Regarding competition 
in biology, Darwin \cite{Darwin1968} wrote the following:

\begin{quote}
\begin{em}
[\ldots] it is the most closely allied forms -- varieties of the same species 
and species of the same genus or of related genera -- which, from having nearly 
the same structure, constitution, and habits, generally come into the severest 
competition with each other.
\end{em}
\end{quote}

\noindent Likewise, the words in a synset, having nearly the same meaning, 
generally come into the severest competition with each other.

The project of understanding how synsets change their leaders raises
a number of questions: How much change is due to random events and how much 
is due to sustained pressures? What are the features of a word that 
determine its fitness for survival and growth in frequency? Is it possible 
to predict the outcome of a struggle for dominance of a synset? 

We present an algorithm that uses supervised learning to predict the 
leading member of a synset, applying features based on a word's length, 
its letters, and its corpus statistics. The algorithm gives insight 
into which features cause a synset's leader to change. 

The algorithm is evaluated with a dataset of 4,000 WordNet synsets, 
containing 9,000 English words and their frequencies in GBNC from 
1800 AD to 2000 AD. The dataset enables us to study how English words 
have evolved over the last two hundred years. In this period, more than 
42\% of the 4,000 synsets had at least one change in leader. In a typical 
fifty-year interval, about 16.5\% of the synsets experience a change in 
leader. The algorithm can predict leadership changes (including the 
identity of the new leader) fifty years ahead with an F-score of 
38.5--43.3\%, whereas random guessing yields an F-score of 17.3--24.8\%. 

The main contributions of this paper are (1) the creation and release
of a dataset of 4,000 synsets containing a total of 9,000 English words and
their historical frequencies \cite{Turney2019}, (2) a set of features that 
are useful for predicting change in the leader of a synset, (3) software for 
processing the dataset with supervised learning \cite{Turney2019}, generating 
models that can predict changes in synset leadership, (4) a method for 
analysis of the learned models that provides insight into the causes of 
changes in synset leadership.

In the next section, we discuss related work on evolutionary models of
word change. The following section describes how we constructed the
dataset of synsets and provides some statistics about the dataset. 
Next, we present the features that we use to characterize the dataset and 
we outline the learning algorithm. Four sets of experiments are summarized 
in the subsequent section. We then consider limitations and future work and 
present our conclusion.

\section*{Related Work on the Evolution of Words}

Much has been written about the evolution of words. Van Wyhe
\cite{vanWyhe2005} provides a good survey of early research. 
Gray, Greenhill, and Ross \cite{Gray2007} and Pagel 
\cite{Pagel2009} present thorough reviews of recent work. 
Mesoudi \cite{Mesoudi2011} gives an excellent introduction to 
work on the evolution of culture in general. In this section, we 
present a few relevant highlights from the literature on the evolution 
of words.

Darwin believed that his theory of natural selection should be applied 
to the evolution of words \cite{Darwin2003}:

\begin{quote}
\begin{em}
The formation of different languages and of distinct species, and the 
proofs that both have been developed through a gradual process, are 
curiously parallel. [\ldots] The survival or preservation of certain
favoured words in the struggle for existence is natural selection.
\end{em}
\end{quote}

\noindent However, he did not attempt to work out the details
of how words evolve.

Bolinger \cite{Bolinger1953} argued that words with similar 
forms (similar spellings and sounds) should have similar meanings. 
As an example, he gave the words {\em queen} and {\em quean}, the latter
meaning ``a prostitute or promiscuous woman.'' Bolinger claimed 
the word {\em quean} has faded away because it violates his dictum. 

Magnus \cite{Magnus2001} defends the idea that some individual phonemes 
convey semantic qualities, which can be discovered by examining the words 
that contain these phonemes. For example, several words that begin with 
the letter {\em b} share the quality of roundness: {\em bale, ball, bay, 
bead, bell, blimp, blip, blob, blotch, bowl, bulb}. We do not pursue this 
idea here, but we believe resources such as GBNC and WordNet might be 
used to test this intriguing hypothesis.

Petersen et~al. \cite{Petersen2012} find that, as the vocabulary
of a language grows, there is a decrease in the rate at which new words 
are coined. They observe that a language is like a gas that cools as
it expands. Consistent with this hypothesis, we will show that, for 
English, the rate of change in the leadership of synsets has decreased 
over time.

Newberry et~al. \cite{Newberry2017} examine three grammatical changes to 
quantify the strength of natural selection relative to random drift: 
(1) change in the past tense of 36 verbs, (2) the rise of {\em do} in 
negation in Early Modern English, and (3) a sequence of changes in negation 
in Middle English.

Cuskley et~al. \cite{Cuskley2014} study the competition between
regular and irregular verbs. They find that the amount of irregularity
is roughly constant over time, indicating that the pressures to make
verbs conform to rules are balanced by counter-pressures. In our
experiments below, we observe rivalries among various suffixes, 
indicative of similar competing pressures.

Ghanbarnejad et~al. \cite{Ghanbarnejad2014} analyze the dynamics of language 
change, to understand the variety of curves that we see when we plot 
language change over time (as in our Fig~\ref{figure:rapturous--ecstatic}).
They introduce various mathematical models that can be used to gain
a deeper understanding of these curves.

Amato et~al. \cite{Amato2018} consider how linguistic norms evolve
over time. Their aim is to distinguish spontaneous change in norms
from change that is imposed by centralized institutions. They
argue that these different sources of change have distinctive
signatures that can be observed in the statistical data.

As we mentioned in the introduction, several papers consider how words change 
their meanings over time \cite{Mihalcea2012,Mitra2015,Xu2015,Hamilton2016}. 
For example, Mihalcea and Nastase \cite{Mihalcea2012} discuss the shift in 
meaning of {\em gay}, from expressing an emotion to specifying a sexual 
orientation. Instead of studying how the meaning of a word shifts over
time (same word, new meaning), we study how the the most frequent word in 
a synset shifts over time (same meaning, new word). As Darwin 
\cite{Darwin2003} put it, we seek to understand the ``preservation of 
certain favoured words in the struggle for existence.''

\section*{Building Datasets of Competing Words}

Predicting the rise and fall of words in a synset could be viewed as a
time series prediction problem, but we prefer another point of view. 
Fifty years from now, will {\em ecstatic} still dominate its synset, or 
will it perhaps be replaced by {\em rapt}? This is a classification 
problem, rather than a time series prediction problem. The classes are 
{\em winner} and {\em loser}. 

Our algorithm has seven steps. The first four steps involve combining
information from WordNet and GBNC to make an integrated dataset for
studying the competition of words to represent meanings. The last 
three steps involve supervised learning with feature vectors.
We present the first four steps in this section and the last three
steps in the section {\em Learning to Model Word Change}. 

The first four steps yield a dataset that is agnostic about the
feature vectors and algorithms that might be used to analyze the data.
The first step extracts the frequency data we need from GBNC,
the second step sums frequency counts for selected time periods,
the third step groups words into synsets, and the fourth step
splits the data into training and testing sets. The output of
the fourth step is suitable for other researchers to use for 
evaluating their own feature vectors and learning algorithms. 
The results that other researchers obtain with this feature-agnostic 
dataset should be suitable for comparison with our results.

In the section {\em Learning to Model Word Change}, we take as input 
the feature-agnostic dataset that is the output of the fourth step.
The fifth step adds features to the feature-agnostic data, the sixth 
step applies supervised learning, and the seventh step summarizes the 
results.

\subsection*{Past, present, and future}

Imagine that the year is 1950 and we wish to predict which member of the 
{\em rapturous--ecstatic} synset will be dominant in 2000. In principle,
we could use the entire history of the synset up to 1950 to make our
prediction; however, it can be challenging to see a trend in such a 
large quantity of data. 

To simplify the problem, we focus on a subset of the data.
The idea is, to look fifty years into the future, we should look fifty 
years into the past, in order to estimate the pace of change. 
The premise is that this focus will result in a simple, 
easily interpretable model of the evolution of words.

We divide time into three periods, {\em past}, {\em present}, and {\em future}. 
Continuing our example, the {\em past} is 1900, the {\em present} is 1950,
and the {\em future} is 2000. Suppose that data from these three periods
constitutes our testing dataset. We construct the training set by shifting 
time backwards by fifty years, relative to the testing dataset. In the 
training dataset, the {\em past} is 1850, the {\em present} is 1900, and 
the {\em future} is 1950. This lets us train the supervised classification 
system without peeking into our supposed future (the year 2000, as seen 
from 1950).

\subsection*{Integrating GBNC with WordNet}

All words in WordNet are labeled with sense information. GBNC includes part 
of speech information, but it does not have word sense information. To 
bridge the word sense gap between GBNC and WordNet, we have chosen to 
restrict our datasets to the monosemous (single-sense) words in WordNet. 
A WordNet synset is included in our dataset only when every word in the 
synset is monosemous. 

For example, {\em rapt} is represented in GBNC as {\em rapt\_ADJ}, meaning
{\em the adjective rapt}. We map the GBNC frequency count for {\em rapt\_ADJ}
to the WordNet representation {\em rapt\#a\#1}, meaning {\em the first
sense of the adjective rapt}. We can do this because the adjective {\em rapt}
has only one possible meaning, according to WordNet. If {\em rapt\_ADJ} had two 
senses in WordNet, {\em rapt\#a\#1} and {\em rapt\#a\#2}, then we would not 
know how to properly divide the frequency count of {\em rapt\_ADJ} 
in GBNC over the two WordNet senses, {\em rapt\#a\#1} and {\em rapt\#a\#2}. 

The frequency counts in GBNC for the five words {\em ecstatic\_ADJ}, 
{\em enraptured\_ADJ}, {\em rapt\_ADJ}, {\em rapturous\_ADJ}, and 
{\em rhapsodic\_ADJ} are mapped to the five word senses {\em ecstatic\#a\#1}, 
{\em enraptured\#a\#1}, {\em rapt\#a\#1}, {\em rapturous\#a\#1}, and 
{\em rhapsodic\#a\#1} in the {\em rapturous--ecstatic} synset in WordNet. This 
is permitted because {\em ecstatic\_ADJ}, {\em enraptured\_ADJ}, {\em rapt\_ADJ}, 
{\em rapturous\_ADJ}, and {\em rhapsodic\_ADJ} are all monosemous in WordNet. 

The word {\em enraptured} could be either an adjective or the past tense 
of a verb. However, it is not ambiguous when it is tagged with a part
of speech, {\em enraptured\_ADJ} or {\em enraptured\_VERB}. Thus we can
map the frequency count for {\em enraptured\_ADJ} in GBNC to the monosemous 
{\em enraptured\#a\#1} in WordNet.

There are other possible ways to bridge the word sense gap between GBNC 
and WordNet. We will discuss this in the section on future work.

\subsection*{Potential Limitations of GBNC and WordNet}

Before we explain how we combine GBNC and WordNet, we should
discuss some potential issues with these resources. We argue that
the design of our experiments mitigates these limitations.

The corpus we use, {\em English Version 20120701}, contains a relatively
large portion of academic and scientific text \cite{Pechenick2015}. The
word frequencies in this corpus are not representative of colloquial word
usage. However, in our experiments, we only compare relative frequencies 
of words within a synset. The bias towards scientific text may affect 
a synset as a whole, but it is not likely to affect relative frequencies 
within a synset, especially since we restrict our study to monosemous 
words. The benefit of using {\em English Version 20120701}, compared 
to other more colloquial corpora, is its large size, which enables 
greater coverage of WordNet words and more robust statistical analysis.

Monosemous words tend to have lower frequencies than polysemous words,
thus restricting the study to monosemous words creates a bias towards
lower frequency words. This could have a quantitative impact on our
results. For example, lower frequency words may change more rapidly
than higher frequency words, so the rate of change that we see with
monosemous words may not be representative of what we would see with
polysemous words. Even though this may have an impact on the specific 
numerical values we report, the same evolutionary mechanisms apply to 
both the monosemous and polysemous words, and thus the broader 
conclusions on the trends reported here should be common to both.

In this study, we assume that WordNet synsets are stable over the period 
of time we consider, 1800 to 2000. Perc \cite{Perc2012} argues that
English evolved rapidly from 1520 to 1800 and then slowed down from
1800 to 2000. Although it is possible that WordNet synsets may be missing 
some words that were common around 1800, WordNet appears to have good 
coverage for the last 200 years. Inspection of WordNet shows that it 
contains many archaic words that are rarely used today, such as 
{\em palfrey} and {\em paltering}. 

Aside from potentially missing words, another possibility is that
a word may have moved from one synset to another in the last 200 years.
WordNet might indicate that such a word is monosemous and belongs only in 
the later synset. Although this is possible, we do not know of any cases 
where this has happened. It is likely that such cases are relatively 
rare and would have little impact on our conclusions. 

\subsection*{Building datasets}

We build our datasets for studying the evolution of words in four steps,
as follows.

{\bf Step~1:} {\em Extract WordNet unigrams from GBNC.} For the first sense
of each unigram word in WordNet, if it contains only lower case letters
and has at least three letters (for example, {\em ecstatic\#a\#1}),
then we look for the corresponding word in GBNC ({\em ecstatic\_ADJ})
and find its frequency for each year. GBNC records both the number of tokens 
of a word and the number of books that contain a word. By {\em frequency},
we mean the number of tokens. In Step~3, when we group words into synsets, 
we will eliminate synsets that contain words with a second sense (that is, 
words that are not monosemous). We believe that, if a synset contains a word 
that is not monosemous, then it is best to avoid the whole synset, rather 
than merely removing the polysemous word from the synset. 

{\bf Step~2:} {\em Sum the frequency counts for selected time periods.} 
GBNC has data extending from 1500 AD to 2008 AD, but the data is sparse
before 1800. We sample GBNC for frequency information every fifty years 
from 1800 to 2000. To smooth the data, we take the sum of the frequency 
counts over an eleven-year interval. For example, for the year 1800, we 
take the sum of the frequency counts from 1795 to 1805; that is, 
1800 $\pm$ 5. Frequency information from 1806 to 1844, from 1856 
to 1894, and so on, is not used. (See the first column of 
Table~\ref{table:time-periods}.)

{\bf Step~3:} {\em Group words into synsets.} Each synset must contain at
least two words; otherwise there is no competition between words.
Every word in a synset must be monosemous. If any word in a given synset
has two or more senses, the entire synset is discarded.

{\bf Step~4:} {\em Split the data into training and testing sets.}
Each training or testing set covers exactly three time periods: {\em past}, 
{\em present}, and {\em future}. Each training set is shifted fifty years
backward from its corresponding testing set. Given a sampling cycle of fifty 
years, from 1800 to 2000, we have two train--test pairs, as shown in 
Table~\ref{table:time-periods}. We remove a synset from a training or 
testing set if there is a tie for first place in the {\em present} or 
the {\em future}. We also remove a synset from a training or testing set if
it contains any words that are unknown in the {\em present}. A word is 
considered to be unknown in the {\em present} if it has a frequency of zero 
in the {\em present}. If a word’s frequency in the {\em present} 
is zero, then it is effectively dead and it is not a serious candidate for 
being a future winner. We decided that including synsets that contain dead 
words would artificially inflate the algorithm's score.

\begin{table}[!ht]
\begin{center}
\begin{tabular}{lrrrr}
\hline
Period           & Train1  & Test1   & Train2  & Test2 \\
\hline
1800 $\pm$ 5     & past    &         &         & \\
1850 $\pm$ 5     & present & past    & past    & \\
1900 $\pm$ 5     & future  & present & present & past \\
1950 $\pm$ 5     &         & future  & future  & present \\
2000 $\pm$ 5     &         &         &         & future \\
\hline
Synsets          & 2,528   & 3,484   & 3,484   & 4,092 \\
Words            & 5,640   & 7,795   & 7,795   & 9,198 \\
Words per synset & 2.23    & 2.24    & 2.24    & 2.25 \\
Change           & 17.3\%  & 19.0\%  & 19.0\%  & 13.3\% \\
\hline
\end{tabular}
\end{center}
\caption{\label{table:time-periods} Time periods for the training and 
testing sets, given a fifty-year cycle of eleven-year samples.
The average synset contains 2.23 to 2.25 words.}
\end{table}

The bottom rows of Table~\ref{table:time-periods} give some
summary statistics. {\em Synsets} is the number of
synsets in each dataset and {\em words} is the number of words.
{\em Change} is the percentage of synsets where the leader changed 
between the {\em present} and the {\em future}.

Table~\ref{table:rapturous--ecstatic} shows a sample of the output
of Step~4. The sample is the entry for the {\em rapturous--ecstatic} 
synset in the Test1 dataset. In 1850 (considered to be the past in Test1), 
{\em rapturous} was the leading member of the synset. In 1900 and 1950 
(considered to be the present and the future in Test1), {\em ecstatic} 
took over the leadership.

\begin{table}[!ht]
\begin{center}
\begin{tabular}{lrrr}
\hline
Test1 dataset      & Past frequency & Present frequency & Future frequency \\
Ecstatic synset    & 1850 $\pm$ 5   & 1900 $\pm$ 5      & 1950 $\pm$ 5 \\
\hline
ecstatic\#a\#1     &      5,576     & {\bf 21,716}      & {\bf 30,829} \\
enraptured\#a\#1   &      4,334     &       7,148       &       5,263 \\
rapt\#a\#1         &      5,243     &      18,750       &      14,845 \\
rapturous\#a\#1    & {\bf 8,645}    &      15,320       &       9,544 \\
rhapsodic\#a\#1    &         45     &         696       &       3,595 \\
\hline
\end{tabular}
\end{center}
\caption{\label{table:rapturous--ecstatic} A sample of the Test1
dataset entries for the {\em rapturous--ecstatic} synset. The highest
frequencies for each time period are marked in bold, indicating the winners.}
\end{table}

\subsection*{The amount of change in the datasets}

A key question about how language evolves is how frequently the 
meanings have new leaders, and how this rate of leadership change 
itself changes over time. Table~\ref{table:changes} summarizes
the amount of change, given a cycle of fifty years from 1800 to 
2000, with word frequency counts summed over eleven-year intervals. 
Here we analyze the data after Step~3 and before Step~4. The table 
shows that 42\% of the synsets had at least one change of leadership over
the course of 200 years. Since only five periods are sampled (see the first 
column in Table~\ref{table:time-periods}), at most four changes are possible. 
The bottom row of Table~\ref{table:changes} shows that two synsets 
experienced this maximum level of churn.  

\begin{table}[!ht]
\begin{center}
\begin{tabular}{lrr}
\hline
$\ge$ {\em N} changes &   Number of &  Percent of \\
                      &     synsets &     synsets \\
\hline
$\ge$ 1 change        &       1,817 &      42.14\% \\
$\ge$ 2 changes       &         518 &      12.01\% \\
$\ge$ 3 changes       &          65 &       1.51\% \\
$=$ 4 changes         &           2 &       0.05\% \\
\hline
\end{tabular}
\end{center}
\caption{\label{table:changes} The frequency of synset leadership changes 
over 200 years, given a fifty-year cycle of eleven-year samples. Change of
leadership is common.}
\end{table}

The amount of change that we see depends on the cycle length (fifty years)
and the interval for smoothing (eleven years). Shorter cycles and shorter smoothing
intervals will show more change, but we should also expect to see more random 
noise. In the setup we have described in this section, we chose
relatively long cycles and smoothing intervals, in an effort to minimize noise.
By summing over eleven-year intervals and sampling over 
fifty-year intervals, we greatly reduce the risk of detecting random synset 
changes. On the other hand, we increase the risk that we are missing true 
synset changes. In our experiments, we will explore different cycle lengths. 

\section*{Learning to Model Word Change}

Now that we have training and testing datasets, we apply supervised 
learning to predict when the leadership of a synset will change. We do 
this in three more steps, as follows.

{\bf Step~5:} {\em Generate feature vectors for each word.} We describe
how we generate feature vectors for each word in the next section.
Our final aim is to make predictions at the level of synsets. For example, given 
the {\em past} and {\em present} data for the {\em rapturous--ecstatic} synset
in Test1 (see Table~\ref{table:rapturous--ecstatic}), we want to
predict that {\em ecstatic} will be the leader of the synset in the 
{\em future}. To make such predictions, we first work at the level of
individual words, then we later move up to the synset level. 

{\bf Step~6:} {\em Train and test a supervised learning system at the word level.}
For each word, we need a model that can estimate its future fitness;
that is, the number of tokens the word will have in the future, relative to
its competition (the other words in the synset). We treat this as a binary
classification task, where the two classes are {\em winner} and {\em loser}.
However, for Step~7, we need to estimate the probability of being
a winner, rather than simply guessing the class. Later we will
explain how we obtain probabilities. The probability of winning can be
interpreted as the estimated future fitness of a word.

{\bf Step~7:} {\em Summarize the results at the synset level.} Given 
probabilities for each of the words in the synset, we guess the winner
by simply selecting the word with the highest probability of winning.
Thus, for each synset, we have one final output: the member of the synset
that we expect to be the winner. The probabilities described in Step~6
are more useful for this step than the binary classes, {\em winner} and 
{\em loser}. Probabilities are unlikely to yield ties, whereas binary 
classes could easily yield two or more winners or zero winners.

\subsection*{Feature vectors for words}

We represent each word with a vector consisting of eight features and the 
target class. There are two length-based features, three character-based 
features, and four corpus-based elements (three features and the class). 
We will first define the features, then give examples of the vectors.

{\bf Feature~1:} {\em Normalized length} is the number of characters in 
the given word, divided by the maximum number of characters for any word 
in the given synset. The idea is that shorter words might be more fit, 
since they can be generated with less effort. [{\em length-based, real-valued}]

{\bf Feature~2:} {\em Syllable count} is the number of syllables in the 
given word \cite{Bowers2016}. The intuition behind this feature is that 
{\em normalized length} applies best to written words, whereas {\em syllable 
count} applies best to spoken words, so the two features may be complementary. 
[{\em length-based, integer-valued}]

{\bf Feature~3:} {\em Unique ngrams} is the set of letter trigrams in the 
given word that are not shared with any other words in the given synset. 
This is not a single feature; it is represented by a high-dimensional 
sparse binary vector. The motivation for this feature vector is that there 
may be certain trigrams that enhance the fitness of a word. Before we split 
the given word into trigrams, we add a vertical bar to the beginning and ending 
of the word, so that prefix and suffix trigrams are distinct from interior 
trigrams. For example, {\em ecstatic} becomes {\em $\vert$ecstatic$\vert$}, 
which yields the trigrams {\em $\vert$ec}, {\em ecs}, {\em cst}, {\em sta}, 
{\em tat}, {\em ati}, {\em tic}, and {\em ic$\vert$}. However, the trigram
{\em ic$\vert$} is not unique to {\em ecstatic}, since it is shared with 
{\em rhapsodic}. Likewise, {\em rapturous} shares its first four letters
with {\em rapt}, so the unique trigrams for {\em rapturous} must omit
{\em $\vert$ra}, {\em rap}, and {\em apt}. Also, overlap with {\em enraptured}
means that {\em ptu} and {\em tur} are not unique to {\em rapturous}. The reason 
for removing shared trigrams is that they cannot distinguish the winner from a 
loser; we want to focus on the features that are unique to the winner. 
[{\em character-based, sparse binary vector}]

{\bf Feature~4:} {\em Shared ngrams} is the fraction of letter trigrams 
in the given word that are shared with other words in the given synset. 
A large fraction indicates that the given word is quite similar to its 
competitors, which might be either beneficial or harmful for the word. 
[{\em character-based, real-valued}]

{\bf Feature~5:} {\em Categorial variations} is the number of categorial 
variations of the given word. One word is considered to be a {\em categorial 
variation} of another word when the one word has been derived from the other. 
Often, but not always, the two words have different parts of speech.
For example, {\em hunger\_NOUN}, {\em hunger\_VERB}, and {\em hungry\_ADJ} are
categorial variations of each other \cite{Habash2003,Catvar2003}.
The calculation of categorial variations takes the birth date of a word into
account; that is, the number of categorial variations of a word does not 
include variations that were unknown at the specified {\em present} time.
A word with many categorial variations is analogous to a species with
many similar species in its genus. This suggests that the ancestor of the species 
was highly successful \cite{Hunt2007}. [{\em character-based, integer-valued}]

{\bf Feature~6:} {\em Relative growth} is the growth of a word {\em relative} 
to its synset. Suppose the word {\em ecstatic} occurs $n$ times in the 
{\em present}; that is, $n$ is the {\em raw frequency} of {\em ecstatic} in the 
{\em present}. Suppose the total of the raw frequencies of the six words in 
the {\em rapturous--ecstatic} synset is $N$. The {\em relative frequency} of 
{\em ecstatic} is $n/N$, the frequency of {\em ecstatic} relative to its synset. 
Let $f_1$, $f_2$, and $f_3$ be a word's relative frequencies in the {\em past}, 
{\em present}, and {\em future}, respectively. Let $\Delta$ be {\em relative 
growth}, the change in relative frequency from {\em past} to {\em present}, 
$\Delta = f_2 - f_1$. The {\em relative growth} of {\em ecstatic} is its 
relative frequency in the {\em present} minus its relative frequency in the 
{\em past}. If the synset as a whole is declining, the word in the synset that 
is declining most slowly will be growing {\em relative} to its synset. 
[{\em corpus-based, real-valued}]

{\bf Feature~7:} {\em Linear extrapolation} is the expected relative frequency 
of the given word in the {\em future}, calculated by linear extrapolation 
from the relative frequency in the {\em past} and the {\em present}. 
Since the time interval from {\em past} to {\em present} 
is the same as the time interval from {\em present} to {\em future} (fifty years), 
linear extrapolation leads us to expect the same amount of change from the
{\em present} to the {\em future}; that is, $f_3 = f_2 + \Delta = 2 f_2 - f_1$. 
[{\em corpus-based, real-valued}]

{\bf Feature~8:} {\em Present age} is the age of the given word, relative 
to the {\em present}. We look in GBNC for the first year in which the 
given word has a nonzero frequency, and we take this year to be the 
birth year of the word. We then subtract the birth year from
the {\em present} year, where the {\em present} year depends on the given 
dataset. In Step~4, we require all words to have nonzero frequencies in the
{\em present}, so the birth year of a word is necessarily before the 
{\em present} year. The idea behind this feature is that older words should be 
more stable. [{\em corpus-based, integer-valued}]

{\bf Target class:} The {\em target class} has the value 1 ({\em winner}) if the 
given word has the highest frequency in the given synset in the {\em future}, 
0 ({\em loser}) otherwise. Ties were removed in Step~4, thus exactly one word 
in the given synset can have the value 1 for this feature. The {\em class} is the 
only element in the vector that uses {\em future} data, and it is only visible to 
the learning algorithm during training. The time period that is the {\em future} 
in the training data is the {\em present} in the testing data (see 
Table~\ref{table:time-periods}). [{\em corpus-based, binary-valued}]

Table~\ref{table:features} shows a sample of the output of Step~5. The sample
displays the values of the elements in the vectors for {\em rapturous} and 
{\em ecstatic} in the Test1 dataset. {\em Unique ngrams} is actually a vector 
with 3,660 boolean dimensions. Each dimension corresponds to a trigram. The four 
trigrams that we see for {\em rapturous} in Table~\ref{table:features} have 
their values set to 1 in the high-dimensional boolean vector. The remaining 
3,656 trigrams have their values set to 0.

\begin{table}[!ht]
\begin{center}
\begin{tabular}{lrr}
\hline
Feature               &  rapturous\#a\#1 &  ecstatic\#a\#1 \\
\hline
Normalized length     &            0.900 &          0.800 \\
Syllable count        &                3 &              3 \\
Unique ngrams         & uro, rou, ous, us$\vert$ 
& $\vert$ec, ecs, cst,  sta, tat, ati, tic \\
Shared ngrams         &            0.556 &          0.125 \\ 
Categorial variations &                3 &              2 \\
Relative growth       &         $-$0.122 &          0.107 \\
Linear extrapolation  &            0.119 &          0.449 \\
Present age           &              258 &            213 \\
Target class          &                0 &              1 \\
\hline
\end{tabular}
\end{center}
\caption{\label{table:features} A sample of the Test1 vector elements for 
two of the five words in the {\em rapturous--ecstatic} synset.}
\end{table}

\subsection*{Supervised learning of probabilities}

We use the naive Bayes classifier \cite{John1995} in Weka 
\cite{Witten2016,Weka2015} to process the datasets. Naive Bayes 
estimates the probabilities for the target class by applying Bayes' 
theorem with the assumption that the features are independent. We 
chose the naive Bayes classifier because it is fast, robust, it handles
a variety of feature types, and the output model is easily interpretable.

The naive Bayes classifier in Weka has a number of options. We used
the default settings, which apply normal (Gaussian) distributions to 
estimate probabilities. The data is split into two parts, feature vectors 
for which the {\em class} is 1 and feature vectors for which the 
{\em class} is 0. Each feature is then modeled by its mean and 
variance for each {\em class} value, assuming a Gaussian distribution.
That is, we have two Gaussians for each feature.

\section*{Experiments with Modeling Change}

This section presents four sets of experiments. The first experiment evaluates
the system as described above; we call this system NBCP (Naive Bayes Change 
Prediction). The second experiment evaluates the impact of removing features 
from NBCP to discover which features are most useful. The third experiment 
varies the cycle length from thirty years to sixty years. The final experiment 
takes a close look at the model that is induced by the naive Bayes classifier, 
in an effort to understand what it has learned.

\subsection*{Experiments with NBCP}

Table~\ref{table:time-periods} tells us that 19.0\% of the synsets in Test1 
and 13.3\% of the synsets in Test2 undergo a change of leadership. In datasets
like this, where there is a large imbalance in the classes (81.0--86.7\%
in class~0 versus 13.3--19.0\% in class~1), accuracy is not the appropriate
measure of system performance. We are particularly interested in synsets where 
there is a change of leadership, but these synsets form a relatively small
minority. Therefore, as our performance measures, we use precision, recall, 
and F-score for leadership change, as explained in Table~\ref{table:contingency}.

\begin{table}[!ht]
\begin{center}
\begin{tabular}{|c|c|c|}
\cline{2-3}
\multicolumn{1}{c|}{} & Condition                 & Condition  \\
\multicolumn{1}{c|}{} & positive                  & negative \\
\hline
Predicted             & True positive ({\em tp})  & False positive ({\em fp}) \\
positive              & changed \& right          & stable \& wrong \\
\hline
Predicted             & False negative ({\em fn}) & True negative ({\em tn})\\
negative              & changed \& wrong          & stable \& right \\
\hline
\end{tabular}
\end{center}
\caption{\label{table:contingency} The $2 \times 2$ contingency table
for change in the leadership of a synset.}
\end{table}

The term {\em changed} in Table~\ref{table:contingency} means that the 
{\em present} leader of the given synset is different from the 
{\em future} leader of the synset, whereas {\em stable}
means that the {\em present} and {\em future} leader are the same. By
{\em right}, we mean that the given algorithm correctly predicted the
{\em future} leader of the synset, whereas {\em wrong} means that the
given algorithm predicted incorrectly. True positive, {\em tp}, is the
number of synsets that experienced a change in leadership ({\em changed})
and the given algorithm correctly predicted the new leader ({\em right}).
The other terms, {\em fp}, {\em fn}, and {\em tn}, are defined analogously,
by their cells in Table~\ref{table:contingency}.

Now that we have the definitions of {\em tp}, {\em fp}, {\em fn}, and 
{\em tn} in Table~\ref{table:contingency}, we can define {\em precision}, 
{\em recall}, and \mbox{{\em F-score}} \cite{vanRijsbergen1979,Lewis1995}:

\begin{align}
\textrm{precision} &= \frac{tp}{tp + fp} \\
\textrm{recall}    &= \frac{tp}{tp + fn} \\
\textrm{F-score}   &= 2 \cdot \frac{
  \textrm{precision} \cdot \textrm{recall}
}{
  \textrm{precision} + \textrm{recall}
}
\end{align}

\noindent The \mbox{F-score} is the harmonic mean of precision and 
recall. For all three of the above equations, we use the convention that 
division by zero yields zero. The trivial algorithm that guesses there 
is never a change in leadership will have a {\em tp} count of zero, and 
therefore a precision, recall, and \mbox{F-score} of zero. On the other 
hand, the accuracy of this trivial algorithm would be 81.0--86.7\%, which
illustrates why accuracy is not appropriate here.

Table~\ref{table:NBCP} shows the performance of the NBCP system
on the two testing sets. With 3,484 synsets, the 95\% confidence 
interval for the scores is $\pm$ 1.6\%, calculated using the Wilson 
score interval \cite{Newcombe1998}; thus the F-score for the NBCP 
system (38.5--43.3\%) is significantly better than random guessing
(17.3--24.8\%), due to the much higher precision of the NBCP system, 
which compensates for the lower recall of NBCP, compared to random.

\begin{table}[!ht]
\begin{center}
\begin{tabular}{lrr}
\hline
Statistic              &   Test1 &   Test2 \\
\hline
Number of synsets      &   3,484 &   4,092 \\
Percent changed        &    19.0 &    13.3 \\
Percent stable         &    81.0 &    86.7 \\
\hline
Precision for random   &    16.9 &    10.9 \\
Recall for random      &    46.1 &    42.4 \\
F-score for random     &    24.8 &    17.3 \\
\hline
Precision for NBCP     &    51.0 &    47.3 \\
Recall for NBCP        &    31.0 &    40.0 \\
F-score for NBCP       &    38.5 &    43.3 \\
\hline
\end{tabular}
\end{center}
\caption{\label{table:NBCP} Various statistics for NBCP and
random systems. All numbers are percentages, except for {\em number of 
synsets}.}
\end{table}

In Table~\ref{table:NBCP}, by {\em random}, we mean an algorithm that 
simulates probabilities by randomly selecting a real number from the 
uniform distribution over the range from zero to one. In Step~6, 
probabilities are calculated at the level of individual words, not at 
the level of synsets. Consider the synset \{{\em abuzz}, {\em buzzing}\}. 
The naive Bayes algorithm treats each of these words independently. 
When it considers {\em abuzz}, it does not know that {\em buzzing} 
is the only other choice. Therefore, when it assigns a probability to 
{\em abuzz} and another probability to {\em buzzing}, it makes no 
effort to ensure that the sum of these two probabilities is one. It 
only ensures that the probability of {\em abuzz} being a {\em winner} 
in the future plus the probability of {\em abuzz} being a {\em loser} 
in the future equals one. Our random system follows the same approach. 
For {\em abuzz}, it randomly selects a number from the uniform 
distribution over the range from zero to one, and this number is taken 
as the probability that {\em abuzz} will be the winner. For {\em buzzing}, 
it randomly selects another number from the uniform distribution over 
the range from zero to one, and this is the probability that {\em buzzing} 
will be the winner. There is no attempt to ensure that these two simulated
probabilities sum to one. 

\subsection*{Feature ablation studies}

Table~\ref{table:ablation1} presents the effect of removing a single
feature from NBCP. The numbers report the \mbox{F-score} when a feature 
is removed minus the \mbox{F-score} with all features present. If every 
feature is contributing to the performance of the system, then we expect 
to see only negative numbers; removing any feature should reduce performance. 
Instead, we see positive numbers for {\em syllable count} and {\em unique ngrams}, 
but these positive numbers are not statistically significant.

\begin{table}[!ht]
\begin{center}
\begin{tabular}{lrr}
\hline
Feature               &          Test1  &          Test2  \\
\hline
Normalized length     &           0.00  &        $-$0.61  \\
Syllable count        &           0.12  &           0.03  \\
Unique ngrams         &   {\bf $-$3.49} &           0.71  \\
Shared ngrams         &           0.00  &           0.00  \\
Categorial variations &        $-$0.07  &        $-$0.58  \\
Relative growth       &        $-$1.43  &        $-$0.76  \\
Linear extrapolation  &   {\bf $-$2.54} &  {\bf $-$10.08} \\
Present age           &        $-$0.19  &        $-$0.29  \\
\hline
\end{tabular}
\end{center}
\caption{\label{table:ablation1} The drop in \mbox{F-score}
when a feature is removed from the NBCP system. Numbers 
that are statistically significant with 95\% confidence
are marked in bold. Negative numbers indicate that a feature
is making a useful contribution to the system.}
\end{table}

The numbers in Table~\ref{table:ablation2} report the \mbox{F-score}
for each feature alone minus the F-score for random guessing. We expect 
only positive numbers, assuming every feature is useful, but there is 
one signficantly negative number, for {\em shared ngrams} in Test1. 

\begin{table}[!ht]
\begin{center}
\begin{tabular}{lrrrr}
\hline
Feature               &         Test1  &         Test2  \\
\hline
Normalized length     &     {\bf 6.36} &     {\bf 6.85} \\
Syllable count        &          0.86  &     {\bf 3.48} \\
Unique ngrams         &     {\bf 1.66} &     {\bf 3.71} \\
Shared ngrams         &  {\bf $-$2.72} &       $-$0.08  \\
Categorial variations &       $-$0.02  &          1.22  \\
Relative growth       &     {\bf 3.95} &    {\bf 10.10} \\
Linear extrapolation  &     {\bf 9.62} &   {\bf  23.49} \\
Present age           &     {\bf 5.01} &     {\bf 5.52} \\
\hline
\end{tabular}
\end{center}
\caption{\label{table:ablation2} The \mbox{F-score} of each feature
alone minus the \mbox{F-score} of random guessing. Numbers 
that are statistically significant with 95\% confidence
are marked in bold. Positive numbers indicate that a feature
is better than random guessing.}
\end{table}

Table~\ref{table:ablation2} shows that {\em linear extrapolation} is
the most powerful feature. Comparing Table~\ref{table:ablation1} with 
Table~\ref{table:ablation2}, we can see that the features mostly do 
useful work (Table~\ref{table:ablation2}), but their contribtion is hidden 
when the features are combined (Table~\ref{table:ablation1}). The 
comparison tells us that the features are highly correlated with each other. 

\subsection*{Experiments with varying time periods}

The NBCP system samples GBNC with a cycle of fifty years, as described 
above and shown in Table~\ref{table:time-periods}. In this section, we 
experiment with cycles from thirty years up to sixty years. 
Table~\ref{table:varying-time} reports the \mbox{F-scores} for
the different cycle times. The dates given are for the 
{\em future} period of each testing dataset, since that is the
target period for our predictions.

\begin{table}[!ht]
\begin{center}
\begin{tabular}{lrrrr}
\hline
Cycle              &        Test1 &        Test2 &        Test3 &        Test4 \\
\hline
30 years           & 1910 $\pm$ 5 & 1940 $\pm$ 5 & 1970 $\pm$ 5 & 2000 $\pm$ 5 \\
F-score for NBCP   &         34.4 &         40.6 &         38.4 &         38.8 \\
F-score for random &         21.0 &         18.0 &         15.5 &         12.6 \\
\hline
40 years           & 1920 $\pm$ 5 & 1960 $\pm$ 5 & 2000 $\pm$ 5 \\
F-score for NBCP   &         34.7 &         38.3 &         42.5 \\
F-score for random &         22.7 &         19.0 &         16.1 \\
\cline{1-4}
50 years           & 1950 $\pm$ 5 & 2000 $\pm$ 5 \\
F-score for NBCP   &         38.5 &         43.3 \\
F-score for random &         24.8 &         17.3 \\
\cline{1-3}
60 years           & 2000 $\pm$ 5 \\
F-score for NBCP   &         39.5 \\
F-score for random &         21.2 \\
\cline{1-2}
\end{tabular}
\end{center}
\caption{\label{table:varying-time} The effect that varying cycle lengths
has on the \mbox{F-score} of NBCP and random guessing.}
\end{table}

We have restricted our date range to the years from 1800 AD to 2000 AD, 
due to the sparsity of GBNC before 1800 AD. We require a minimum of four 
cycles to build one training set and one testing set (see Train1 and 
Test1 in Table~\ref{table:time-periods}). With a sixty-year cycle, the 
four time periods that we use are 1820, 1880, 1940, and 2000. Only the 
final period, 2000, is both a {\em future} period and a {\em testing} 
period. With a seventy-year cycle, the four time periods would be 1790, 
1860, 1930, and 2000. Therefore we prefer not to extend the cycle past 
sixty years. 

Regarding periods shorter than thirty years, the amount of change naturally
decreases as we shorten the cycle period. With less change, prediction could
become more difficult. Table~\ref{table:varying-change} shows how the amount 
of change varies with the cycle period.

\begin{table}[!ht]
\begin{center}
\begin{tabular}{lrrrr}
\hline
Cycle             &        Test1 &        Test2 &        Test3 &        Test4 \\
\hline
30 years          & 1910 $\pm$ 5 & 1940 $\pm$ 5 & 1970 $\pm$ 5 & 2000 $\pm$ 5 \\
Percent changed   &         14.7 &         13.7 &         11.0 &          8.4 \\
Number of synsets &        3,041 &        3,622 &        3,958 &        4,275 \\
\hline
40 years          & 1920 $\pm$ 5 & 1960 $\pm$ 5 & 2000 $\pm$ 5 \\
Percent changed   &         17.5 &         14.5 &         11.0 \\
Number of synsets &        3,038 &        3,732 &        4,203 \\
\cline{1-4}
50 years          & 1950 $\pm$ 5 & 2000 $\pm$ 5 \\
Percent changed   &         19.0 &         13.3 \\
Number of synsets &        3,484 &        4,092 \\
\cline{1-3}
60 years          & 2000 $\pm$ 5 \\
Percent changed   &         15.4 \\
Number of synsets &        3,958 \\
\cline{1-2}
\end{tabular}
\end{center}
\caption{\label{table:varying-change} The effect that varying cycle lengths
has on the percentage of synsets that have changed leadership from {\em present} 
to {\em future}.}
\end{table}

Comparing Table~\ref{table:varying-time} with Table~\ref{table:varying-change},
we see that the F-score of random guessing declines as we approach the
year 2000 (see Table~\ref{table:varying-time}), following approximately the 
same pace as the decline of the percent of changed synsets (see 
Table~\ref{table:varying-change}). On the other hand, the F-score of
NBCP remains relatively steady; it is robust when the percent of changed synsets
varies.

In passing, we note that Table~\ref{table:varying-change} suggests the 
amount of change is decreasing as we approach the year 2000. This confirms 
the analysis of Petersen et al. \cite{Petersen2012}, mentioned in our
discussion of related work.

\subsection*{Interpretation of the learned models}

In this section, we attempt to understand what the learned models 
tell us about the evolution of words. For each feature, the naive Bayes 
classifier generates two Gaussian models, one for class 0 (loser) and
one for class 1 (winner). Because naive Bayes assumes features are 
independent, we can analyze the models for each feature independently. 

Here we are attempting to interpret the trained naive Bayes models,
to gain insight into the role that the various features play in language
change. Since the naive Bayes algorithm assumes the features are independent,
the trained models cannot tell us anything about interactions among
the features. It is likely that there are interesting interactions among 
the features. We leave the study of these interactions, possibly with 
algorithms such as logistic regression, for future work. For now, we 
focus on the individual impact of each feature.

Table~\ref{table:feature-interpretation} shows the means of the
Gaussians (the central peaks of the normal distributions)
for the losers and the winners for each feature. This
table omits {\em unique ngrams}, since it is a high-dimensional
vector, not a single feature. We will analyze {\em unique ngrams}
separately. The table only shows the models for Test1. Test2 follows 
the same general pattern.

\begin{table}[!ht]
\begin{center}
\begin{tabular}{lrrrc}
\hline
Feature               & \multicolumn{3}{c}{Means of Gaussians}  & Mean of the \\
\cline{2-4}
                      &    Losers &   Winners &     Difference  & winners is \ldots \\
\hline
Normalized length     &    0.9128 &    0.9087 &      $-$0.0041  & lower  \\
Syllable count        &    3.2494 &    3.2077 &      $-$0.0417  & lower  \\
Shared ngrams         &    0.4797 &    0.4726 &      $-$0.0071  & lower  \\
Categorial variations &    3.3075 &    3.4432 &         0.1357  & higher \\
Relative growth       & $-$0.0012 &    0.0956 &    {\bf 0.0968} & higher \\
Linear extrapolation  &    0.2058 &    0.8408 &    {\bf 0.6350} & higher \\
Present age           &     130.0 &     180.5 &      {\bf 50.5} & higher \\
\hline
\end{tabular}
\end{center}
\caption{\label{table:feature-interpretation} Analysis of the naive Bayes
models for Test1. The {\em difference} column is the mean of the Gaussian of 
the winners (class 1) minus the mean of the Gaussian of the losers (class 0).
Differences that are statistically significant are marked
in bold. Significance is measured by a two-tailed unpaired {\em t} test 
with a 95\% confidence level. This table omits {\em unique ngrams}, which 
are presented in the next table.}
\end{table}

The two length-based features, {\em normalized length} and {\em syllable
count}, both tend to be lower for winning words. This confirms Bolinger's
\cite{Bolinger1953} view that ``economy of effort'' plays a large
role in the evolution of words; brevity is good. On the other hand,
{\em shared ngrams} also tends to be lower, which implies that we prefer
distinctive words. This sets a limit on brevity, since there is a limited
supply of short words. Brevity is good, so long as words are not too similar.

We mentioned earlier that a word with many {\em categorial variations} is 
analogous to a species with many similar species in its genus, which may 
be a sign of success. This is supported by the naive Bayes model, since the 
winner has a higher mean for {\em categorial variations} than the loser.

The table shows that positive {\em relative growth} is better than negative
{\em relative growth} and a high {\em linear extrapolation} is better than
low, as expected. It also shows a high {\em present age} is good. In life,
we tend to associate age with mortality, but {\em present age} is the age
of a word {\em type}, not a {\em token}; it is analogous to the age of a 
species, not an individual. A species that has lasted for a long time has 
demonstrated its ability to survive.

{\em Unique ngrams} is a vector with 3,660 elements in Test1.
To gain some insight into this vector, we sorted the elements in order
of decreasing absolute difference between the mean of the Gaussian 
for class 0 and the mean for class 1. Table~\ref{table:ngram-interpretation} 
gives the top dozen trigrams with the largest gaps between the means. The 
size of the gap indicates the ability of the trigram to discriminate the 
classes. The {\em difference} column is the mean of the Gaussian of the 
winners (class 1) minus the mean of the Gaussian of the losers (class 0). 
When the difference is positive, the presence of the trigram in a word
suggests that the word might be a winner. When the difference is negative,
the presence of the trigram in a word suggests that the word might be
a loser.

\begin{table}[!ht]
\begin{center}
\begin{tabular}{crrrc}
\hline
Trigrams     & \multicolumn{3}{c}{Means of Gaussians}   & Presence of the \\
\cline{2-4}
             &  Losers   &  Winners  &  Difference      & trigram suggests \ldots \\
\hline
ize          &  0.0055   &  0.0285   &    {\bf 0.0230}  & winner \\
ise          &  0.0289   &  0.0083   & {\bf $-$0.0206}  &  loser \\
nes          &  0.0328   &  0.0134   & {\bf $-$0.0194}  &  loser \\
ty$\vert$    &  0.0112   &  0.0297   &    {\bf 0.0185}  & winner \\
ss$\vert$    &  0.0379   &  0.0202   & {\bf $-$0.0177}  &  loser \\
ity          &  0.0100   &  0.0269   &    {\bf 0.0169}  & winner \\
ze$\vert$    &  0.0022   &  0.0174   &    {\bf 0.0152}  & winner \\
ess          &  0.0373   &  0.0229   & {\bf $-$0.0144}  &  loser \\
se$\vert$    &  0.0206   &  0.0083   & {\bf $-$0.0123}  &  loser \\
lis          &  0.0154   &  0.0032   & {\bf $-$0.0122}  &  loser \\
ic$\vert$    &  0.0228   &  0.0348   &    {\bf 0.0120}  & winner \\
liz          &  0.0022   &  0.0115   &    {\bf 0.0093}  & winner \\
\hline
\end{tabular}
\end{center}
\caption{\label{table:ngram-interpretation} Analysis of the {\em unique
ngrams} features in the naive Bayes models for Test1. The table lists 
the top dozen trigrams with the greatest separation between the means.
Differences that are statistically significant are marked
in bold; all of the differences are significant. Significance is measured 
by a two-tailed unpaired {\em t} test with a 95\% confidence level.}
\end{table}

Before splitting a word into trigrams, we added a vertical bar to the
beginning and end of the word, to distinguish prefix and suffix trigrams
from interior trigrams. Therefore the trigram {\em ty$\vert$} in 
Table~\ref{table:ngram-interpretation} refers to the suffix {\em -ty}.

In the table, we see that a high value for {\em ty$\vert$} or {\em ity}
indicates a winner, but it is better to have low values for {\em nes},
{\em ss$\vert$}, and {\em ess}. Thus the naive Bayes model has confirmed
the conflict between the suffixes {\em -ity} and {\em -ness} 
\cite{Arndt-Lappe2014}: ``Rivalry between the two English nominalising 
suffixes {\em -ity} and {\em -ness} has long been an issue in the 
literature on English word-formation.'' Furthermore, the naive Bayes model 
suggests that {\em -ness} is losing the battle to {\em -ity}. 

We also see that {\em ize}, {\em ze$\vert$}, and {\em liz} are 
indicative of a winner, whereas {\em ise}, {\em se$\vert$} and {\em lis} 
suggest a loser. There is a trend to replace the suffix {\em -ise} with
{\em -ize}. This is known as {\em Oxford spelling}, although it is
commonly believed (incorrectly) that the {\em -ize} suffix is an American 
innovation \cite{Wikipedia2018}.

It is interesting to see that the trigram {\em ic$\vert$} suggests
a winner, and {\em ecstatic} eventually became the leader of the
{\em rapturous--ecstatic} synset (see Fig~\ref{figure:rapturous--ecstatic}).
We looked in the Test1 {\em unique ngrams} vector for {\em ous} and
found that the presence of {\em ous} suggests a loser. This may explain
why {\em ecstatic} eventually won out over {\em rapturous}. However,
we then need to explain the poor performance of {\em rhapsodic} (see 
Fig~\ref{figure:rapturous--ecstatic}). Looking again in the Test1 
{\em unique ngrams} vector, the trigram {\em dic} suggests a loser, 
whereas the trigram {\em tic} suggests a winner.
Although this is consistent with the success of {\em ecstatic} and the
failure of {\em rhapsodic}, it is not clear to us why {\em -tic} should 
be preferred over {\em -dic}, given the apparent similarity of these
suffixes.

\section*{Future Work and Limitations}

Throughout this work, our guiding principle has been simplicity,
based on the assumption that the evolution of words is a complex,
noisy process, requiring a simple, robust approach to modeling.
Therefore we chose a classification-based analysis, instead of a
time series prediction algorithm, and a naive Bayes model, instead
of a more complex model. The success of our approach is encouraging,
and it suggests there is more signal and structure in the data than
we expected. We believe that more sophisticated analyses will reveal
interesting phenomena that our simpler approach has missed.

In particular, there is much room for more features in the feature
vectors for words. We used three types of features: length-based,
character-based, and corpus-based. There are most likely other types
that we have overlooked, and other instances within the three types.
We did a small experiment with phonetic spelling, using the International 
Phonetic Alphabet, but we did not find any benefit. 

As we said in the introduction, the focus of this paper
is {\em selection}. Future work should consider also {\em variation}
and {\em heredity}, the other two components of Darwinian evolution.
There is some past work on predicting variation of words \cite{Skousen1992}.

This general framework may be applicable to other forms of cultural
evolution. For example, the market share of a particular brand within
a specific type of product is analogous to the frequency of a word
within a synset. The fraction of votes for a political party in
a given country is another example.

As we discussed earlier in this paper, to bridge the gap between GBNC and 
WordNet, we restrict our datasets to the monosemous words in WordNet. 
Another strategy would be to allow polysemous words, but map all 
GBNC frequency information for a word to the first sense of the word in 
WordNet. That is, a synset would be allowed to include words that are not
monosemous, but we would assign a frequency count of zero to all senses
other than the first sense. 

WordNet gives a word's senses in order of decreasing frequency 
\cite{Fellbaum1998}. In automatic word sense disambiguation, a standard
baseline is to simply predict the most frequent sense (the first sense) 
for every occurrence of a word. This baseline is difficult to beat 
\cite{Hawker2006}. This could be used as an argument in support of
assigning a frequency count of zero to all senses other than the first 
sense, as a kind of first-order approximation.

We did a small experiment with this strategy, predicting the most 
frequent sense for every occurrence of a word. Our dataset expanded 
from 4,000 synsets containing 9,000 English words to 9,000 synsets 
containing 22,000 English words. The system performance was numerically 
different but qualitatively the same as the results we reported above. 
However, we prefer to take the more conservative approach of only allowing 
monosemous words, since it does not require us to us to assume that 
we can ignore the impact of secondary senses on the evolution of a 
synset. The ideal solution to bridging the gap between GBNC and 
WordNet would be to automatically sense-tag all of the words in GBNC, 
but this would involve a major effort, requiring the cooperation 
of Google.

In the section on related work, we mentioned past research concerned 
with how words change their meanings over time (same word, new meaning) 
\cite{Mihalcea2012,Mitra2015,Xu2015,Hamilton2016}. Let's call this 
{\em meaning-change}. Our focus in this paper has been how meanings change 
their words over time (same meaning, new word; same synset, new leader). 
Let's call this {\em word-change}. These two types of events, 
meaning-change and word-change, are interconnected.

Consider the case of a word that has two possible senses, and thus belongs to 
two different synsets. Suppose that the word's dominant meaning has shifted
over time from the first sense to the second sense, which is a case of
meaning-change. If the frequency of the word in the first synset becomes
sufficiently low and the frequency of the word in the second synset becomes
sufficiently hight, then the word will be less likely to cause a reader 
or listener to be confused when it is used in the second sense. The meaning-change
makes the word less ambiguous and thus it becomes a better candidate for
expressing the meaning of the second synset. This meaning-change may therefore
cause a word-change. The leader of the second synset might be replaced
by the less ambiguous candidate word. 

This example illustrates how meaning-change might cause word-change. We can
also imagine how word-change can cause meaning-change. We expect that future
work will take an integrated approach to these two types of change.
If the words in GBNC were automatically sense-tagged, it would greatly
facilitate this line of research. Here we have alleviated 
the issue of meaning-change and its impact on word-change by restricting 
our dataset to monosemous words. However, we have not completely avoided
the issue of meaning-change, since monosemous words can have shifts in
connotations. There may also be shifts in meanings that WordNet synsets
do not capture. We have avoided precisely defining the relation
between meaning-change and word-change, leaving this for future work,
since there are multiple reasonable ways to define these terms and
their relations.

\section*{Conclusion}

This work demonstrates that change in which word dominates a synset
is predictable to some degree; change is not entirely random. It is
possible to make successful predictions several decades into the future.
Furthermore, it is possible to understand some of the causes of
change in synset leadership.

Of the various features we examined, the most successful is linear
extrapolation. From an evolutionary perspective, this indicates that
there is a relatively constant direction in the natural selection of
words. The same selective pressures are operating over many decades.

We observed that English appears to be cooling; the rate of change is 
decreasing over time. This might be due to a stable environment, as 
suggested by Petersen et al. \cite{Petersen2012}. It might also be 
due to the growing number of English speakers, which could increase 
the inertia of English.

This project is based on a fusion of the Google Books Ngram Corpus
with WordNet. We believe that there is great potential for more
research with this combination of resources. 

This line of research contributes to the sciences of evolutionary 
theory and computational linguistics, but it may also lead to 
practical applications in natural language generation and understanding. 
Evolutionary trends in language are the result of many individuals, 
making many decisions. A model of the natural selection of words can 
help us to understand how such decisions are made, which will enable 
computers to make better decisions about language use.

\section*{Acknowledgments}

We thank the PLOS ONE reviewers for their careful and helpful 
comments on our paper.



\end{document}